\documentclass{article}
\usepackage{ijcai21}

\usepackage{times}
\usepackage{soul}
\usepackage{url}
\usepackage[hidelinks]{hyperref}
\usepackage[utf8]{inputenc}
\usepackage[small]{caption}
\usepackage{graphicx}
\usepackage{amsmath}
\usepackage{amsthm}
\usepackage{booktabs}
\usepackage{algorithm}
\usepackage{algorithmic}
\usepackage{comment}
\usepackage{graphics}
\usepackage{xcolor}
\title{Towards an Interpretable Latent Space in Structured Models \\ for Video Prediction\thanks{\hspace{0.15cm}The paper is accepted at Weakly Supervised Representation Learning Workshop @ IJCAI 2021.}}
\author{
Rushil Gupta$^1$\thanks{equal contribution}
\and
Vishal Sharma$^1$\footnotemark[1]\and
Yash Jain$^{1}$\and\\
Yitao Liang$^2$\and
Guy Van den Broeck$^{2}$\And
Parag Singla$^{1}$
\affiliations
$^1$Indian Institute of Technology Delhi \\
$^2$University of California, Los Angeles\\
\emails
$^1$\{Rushil.Gupta.cs117, vishal.sharma, Yash.Jain.cs517, parags\}@cse.iitd.ac.in

$^2$\{yliang, guyvdb\}@cs.ucla.edu
}

\begin{document}

\maketitle

\begin{abstract}
We focus on the task of future frame prediction in video governed by underlying physical dynamics. We work with models which are object-centric, i.e., explicitly work with object representations, and propagate a loss in the latent space. Specifically, our research builds on recent work by Kipf et al.~\cite{kipf&al20}, which predicts the next state via contrastive learning of object interactions in a latent space using a Graph Neural Network. We argue that injecting explicit inductive bias in the model, in form of general physical laws, can help not only make the model more interpretable, but also improve the overall prediction of model. As a natural by-product, our model can learn feature maps which closely resemble actual object positions in the image, without having any explicit supervision about the object positions at the training time. In comparison with earlier works~\cite{jaques&al20}, which assume a complete knowledge of the dynamics governing the motion in the form of a physics engine, we rely only on the knowledge of general physical laws, such as, world consists of objects, which have position and velocity. We propose an additional decoder based loss in the pixel space, imposed in a curriculum manner, to further refine the latent space predictions. Experiments in multiple different settings demonstrate that while Kipf et al. model is effective at capturing object interactions, our model can be significantly more effective at localising objects, resulting in improved performance in 3 out of 4 domains that we experiment with. Additionally, our model can learn highly intrepretable feature maps, resembling actual object positions.

\end{abstract}

\section{Introduction}
%\begin{comment}
We are interested in the task of future frame prediction in a video given the knowledge about past frames. This task, being important in its own right~\cite{Srivastava&al15}~\cite{Denton&al17}~\cite{Hou&al19}~\cite{Jin&al20}~\cite{Shouno20}~\cite{Oprea&all20}, also has a lot of application in Model-Based Reinforcement Learning~\cite{ohetal15,chiappaetal17,henaffetal17,haetal18,kaiser&al20}.
%~\cite{chiappaetal17},~\cite{henaffetal17},~\cite{haetal18},~\cite{hafneretal19},~\cite{kaiser&al20}. 
A large number of existing work in this direction takes a black-box approach, i.e., given a set of training samples in the form of past and future frame(s), when provided to a neural architecture treated as a black box, the learning algorithm results in a trained network, which can predict the next frame(s) in a video when supplied with preceding frame(s) at inference time. While this approach has met with reasonable success in the literature~\cite{ohetal15}~\cite{haetal18}~\cite{kaiser&al20}, it is faced by limitations such as learning non-intrepretable models, and inability to effectively incorporate domain knowledge in the form of motion based constraints. These approaches are typically trained using a loss in the pixel space, which may not always work well when dealing with objects of varying sizes, such as in Atari-Pong\footnote[1]{https://gym.openai.com/envs/\#atari} where the ball is smaller than the paddles. 

These limitation have resulted in another line of work which takes an {\em object-centric} view of the world, i.e., world is composed of objects, which interact with other, governed by physical  laws~\cite{cheng&al17}~\cite{Kosiorek&al18}~\cite{Hsieh&al18}.
%Our focus in this work is the latter set of approaches adopting the object-centric view. 
Though promising, the research is still in its initial stage with several questions still unanswered. Most existing approaches in this line of work assume some kind of supervised data about objects and relationships between them ~\cite{sunetal18}. This severely limits their applicability to real-life settings when such supervised data may not be easily available. There have also been a set of approaches, which take an unsupervised view of the problem, and directly try to infer the object positions, and their motions, but most of these are trained using the pixel loss in the decoded space. Recent work by Kipf et al.~\cite{kipf&al20} argued that explicitly learning the model by optimizing the loss in latent space has certain advantages of being able to discover smaller objects, and accordingly proposed Constrastive Learning of Structured World Models (CSWM),  a technique which exploits contrastive loss in the latent space, and models object interactions using a Graph Neural Network (GNN) ~\cite{Scarselli&al09}. 

%Another recent work, Physics as Inverse Graphics (PAIG)~\cite{jaques&al20}, 
%A few of the more recent works take an unsupervised view, and try to discover the objects and their interactions, by learning them directly from the data \vscom{Needs citations mentioning related work of cswm and PAIG?}. One such example, is Constrastive Learning of Structured World Models (CSWM)~\cite{kipf&al20}, which models the object dynamics in a latent space using a Graph Neural Network. The learning is done via a novel Contrastive Loss. Though they provide some interpretation of learned feature maps, as possible predictors for objects, there is no direct way impose any additional constraints over this knowledge. Another work is Physics as Inverse Graphics (PAIG)~\cite{jaques&al20} 
%that learns an encoder-decoder architecture capable of estimating physical parameters. PAIG relies on explicit access to differential equations about environment dynamics to estimate physical parameters and coefficients of these equations. This very assumption limits the applicability of the approach since in many settings, such a complete physical description of the motion dynamics may not be available for training.

We argue in this paper, that injecting even simple knowledge of the facts, such as objects having positions, and velocities, which are related by Newtonian laws, can not only make the object-centric approaches more interpretable, but also improve the overall prediction. We build our model on top of the core CSWM~\cite{kipf&al20} architecture. We propose a novel unsupervised method to estimate the position and velocity of objects, while back-propagating the loss computed in the encoded latent space. Our work is different from Physics as Inverse Graphics (PAIG)~\cite{jaques&al20}, 
%A few of the more recent works take an unsupervised view, and try to discover the objects and their interactions, by learning them directly from the data \vscom{Needs citations mentioning related work of cswm and PAIG?}. One such example, is Constrastive Learning of Structured World Models (CSWM)~\cite{kipf&al20}, which models the object dynamics in a latent space using a Graph Neural Network. The learning is done via a novel Contrastive Loss. Though they provide some interpretation of learned feature maps, as possible predictors for objects, there is no direct way impose any additional constraints over this knowledge. Another work is Physics as Inverse Graphics (PAIG)~\cite{jaques&al20} 
that learns an encoder-decoder architecture capable of estimating physical parameters given the knowledge of motion dynamics. PAIG relies on explicit access to differential equations about environment dynamics to estimate physical parameters and coefficients of these equations. This very assumption limits the applicability of the approach since in many settings, such a complete physical description of the motion dynamics may not be available for training. Unlike PAIG, we do not assume the availability of a (complete) physical engine representing motion dynamics but only the generic physical laws governing motion. Our method is {\em unsupervised} in the sense that, we do not assume any knowledge of what the actual objects are, or about their actual positions or velocities.  

Our contributions in this paper can be summarized as follows 1) we introduce a novel unsupervised position loss to estimate position and a set of position constraints that encourage the output channels of (CNN based) extractor to detect one object per channel, 2) we introduce a velocity estimator module guided by an unsupervised velocity loss that encourages the model to satisfy basic relation between position and velocity of objects and, 3) we experiment with using an additional decoder loss in conjunction with the latent space loss, trained in a curriculum manner, to improve the predictions made only using the latent loss.
While in this paper we have added constraints based only on the basic relations among various physical parameters, our long term goal is to propose a method to add various forms of domain knowledge like constraints on velocity of objects or constraints on object interactions.

We evaluate our approach on 4 different environments (i) two different grid-world environments with bouncing shapes but no collision between objects (ii) collision between objects in a grid world, and (iii) 3-body physics. %, and Atari-Pong 
We use Hits@k and Mean Reciprocal Rank (MRR) as evaluation metrics for predictions in the latent space. %We also measure the pixel loss in the decoded image space. 
For Hits@k and MRR, we compare our approach with CSWM, and show that our model is effective at capturing localization of objects, significantly improving the performance of the baseline on 3 out of 4 domains. In our qualitative experiments, we show that our feature maps are much more interpretable, and closely resemble actual object positions in the image, compared to the baseline.
%We also show that our model can predict the object position and velocities to a high degree of accuracy in a varying set of environments.
% Experiments on adding the decoder loss in a training curriculum also shows promising results. 
%~\footnote{We note that PAIG is not applicable, since in our settings, physical dynamics, such as, how the balls bounce back after colliding are not made available to the model.}. 
%We perform an ablation study to show the importance of various losses we have proposed. We also present qualitative analysis depicting long-term prediction of images generated by our image decoder, and the interpretablity of feature maps of our CNN-based extractor.  
The remaining part of the paper is organized as follows, Section 2 discusses the related work, followed by a discussion of the required background in Section 3. Next, we provide description of our approach in Section 4. In Section 5, we explain our experimental setting and present the current results. Section 6 concludes the paper and discusses the future directions.
\section{Related Work}
%We divide our related work broadly into four sub-sections.
%{\bf Video Prediction in Computer Vision:} 
There is a lot work in the Computer Vision community which focuses on the task of video prediction~\cite{Srivastava&al15}~\cite{Denton&al17}~\cite{Hou&al19}~\cite{Jin&al20}~\cite{Shouno20}~\cite{Oprea&all20}. Although related, these works fall along a different line of research since almost all of them use the neural network as a black-box, and propagate loss in the pixel space. This is different from our motivation in this work, where we want to take an object-centric approach, and optimize loss in the latent space, due to various potential advantages, such as ease of interpretation, and better discovery of smaller sized objects. 

Model Based Reinforcement Learning community has built on top of ideas proposed in the vision community, for learning transition models, when the state space is represented by an image ~\cite{ohetal15}~\cite{kaiser&al20}.
%, either as is, or with some alternations, but used as a black box (i.e., without any additional knowledge other than just the input images). 
There is some recent work, which deviates from this approach, and takes an object-centric view, i.e., has an explicit representation of objects, and their attributes such as position. However, many of these works need some kind of supervision about objects, and their features~\cite{sunetal18}, in the absence of which it is difficult to learn. These assumptions limit the applicability of these approaches, since getting such training data may not always be practically possible. Ideally, we would like to be able to learn the model based purely on the input-output images with no supervision on position and velocity.

Some recent works have proposed methods which relax the assumption of supervised data about objects and their features, and work only with input-output image pairs. Though they often work in the latent space representing objects and their properties, in most such cases, the training is based on back-propagation over a loss in the pixel space.This has the potential of giving disproportionate attention to various objects, based on their sizes, which may not always be desirable. For example, in Pong, it may make more sense to detect a ball correctly, although the actual number of pixels that it occupies can be very small. Physics-as-Inverse-Graphics (PAIG)~\cite{jaques&al20} assumes access to a fully differentiable motion dynamics engine, which may be a strong assumption to take for many practical applications.

The closest related work to ours is the Contrastive Learning of Structured World Models (CSWM)~\cite{kipf&al20} which works in an unsupervised setting (i.e., no assumption about knowledge of objects during training), has an explicit representation of objects in terms of feature maps, and fully trains the model using a contrastive loss in the latent space, where object interactions are modeled using a Graph Neural Network~\cite{Scarselli&al09}. The key idea is to back-propagate a loss which measures how similar the next state latent representation is to the one predicted using GNN, while also ensuring that the latent representation of a randomly picked image (negative sample) is far away from the latent representation of the current state. We build on top of this, introducing additional modules, for position and velocity prediction, using knowledge of generic physical laws such as objects exist, and have positions and velocities related by Newtonian laws of motion. Addition of this simple knowledge through our architecture not only makes the model more interpretable by learning feature maps which resemble object positions, but also helps improve overall prediction accuracy in the latent space.

\section{Notations and Background}
We start by defining the notation used in the paper. We assume there are $K$ objects in the scene. The input state at time $t$ is denoted by $s_t$. The corresponding object masks and factorized latent space are denoted by $m_t=(m^k_t)^K_{k=1}$ and $z_t=(z^k_t)^K_{k=1}$ respectively. 
For actuated environments, we denote the action by $a_t$. 
$T(z_t)$ denotes the transition model ($T(z_t, a_t)$ for actuated environments) that predicts the translation in the latent space (denoted by $(\Delta z^k_t)^K_{k=1}$). The predicted next step latent space is denoted by $\hat{z}^k_{t+1}=(z^k_t+\Delta z^k_t)^K_{k=1}$.
The $K$ 2D Gaussian distributions corresponding to $K$ object masks are denoted by $(g^k_t)^K_{k=1}$. The mean of the Gaussian distribution corresponds to position of each object in the mask and is denoted by $p_t=(p^k_t)^K_{k=1}$. The velocities of objects are denoted by $v_t=(v^k_t)^K_{k=1}$. The image predicted by the decoder for time step $t$ is denoted by $\hat{s}_t$.

\subsection{Contrastive Learning of Structured World Models}
The Contrastive Learning of Structured World Models (CSWM)~\cite{kipf&al20} model consists of three parts: 1) CNN based extractor, 2) an object encoder, and 3) Graph Neural Network based transition model. The object extractor takes the current state $s_t$ (or a sequence of previous states in case of environments with inherent velocity) as input and generates $K$ object masks $(m^k_t)^K_{k=1}$, each for an object.
The encoder is an MLP (shared among all objects) that takes object mask $m^k_t$ as input and creates corresponding object encoding $z^k_t$. The $(z^k_t)^K_{k=1}$ form the factorized latent space of the environment. The action space is also assumed to be factorized and for each object the action is represented as a one-hot vector (and all zeros for objects on which no action is applied).
Next, a fully connected graph with $k$ nodes (representing $k$ objects) is created, which is later processed using a Graph Neural Network (GNN)
For each node, the GNN takes as input $z^k_t$ as initial node features along with $a^k_t$ as the corresponding action. The GNN represents the Transition model $T(z_t, a_t)$ and its output represents the translation on the latent encoding of each object denoted by $\Delta z_t=(\Delta z^k_t)^K_{k=1}$.
\begin{equation} \label{cswm:loss}
\begin{split}
L_{CSWM} &= d(z_t +  T(z_t, a_t), z_{t+1}) +  max(0, \gamma - d(\tilde{z}, z_{t+1}))
\end{split}
\end{equation}
The model is trained end-to-end using a contrastive loss between $z_t+\Delta z_t$ and $z_{t+1}$ (equation~\ref{cswm:loss}). Here, $\gamma$ is a hyperparameter for the Hinge-loss.
The loss comprises of two parts, 1) positive part ($d(z_t +  T(z_t, a_t), z_{t+1})$) encourages the encodings $z_t+\Delta z_t$ and $z_{t+1}$ to be close to each other, and 2) negative part ($max(0, \gamma - d(\tilde{z}, z_{t+1}))$) forces the encoding $z_{t+1}$ to be away from encoding of a random negative sample $\tilde{z}$. The distance metric $d(\cdot)$ used is squared Euclidean distance.
%Establish relation b/w position at t t+1 and velocity - add before eqn. 5-------------

\section{Learning Physical Parameters in Structured Latent Space} \label{sec:model}
As discussed in earlier sections, though object masks $(m^k_t)^K_{k=1}$ learned by CSWM can be interpreted to represent locations of objects, there is no explicit estimation of the physical parameters in the model.
Our goal in this section is to describe how we learn a position and a velocity estimator in order to make the latent space interpretable.
We start by defining goals for our model, followed by details on various modules in our architecture.

Given the current state $s_t$ and the previous state $s_{t-1}$, the goal of our model is to learn 1) a set of object masks ($(m^k_t)^K_{k=1}$) each representing an object, 2) a factorized latent space, denoted by $(z^k_t)^K_{k=1}$ that encodes each object, 3) a set of 2D Gaussian distributions ($(g^k_t)^K_{k=1}$) such that the mean of each $g^k_t$ gives the position of the corresponding object, 4) a set of velocity values $(v^k_t)^K_{k=1}$ of each object, 5) a transition model $T(z_t)$ that can predict the encoded next state from the current encoded state and velocity, and 6) an image decoder that is capable of refining the latent space by generating full image of the state from the encoded latent space.

\begin{figure*}[t]
    % \minipage{\linewidth}
      \centering
      \includegraphics[width=\linewidth]{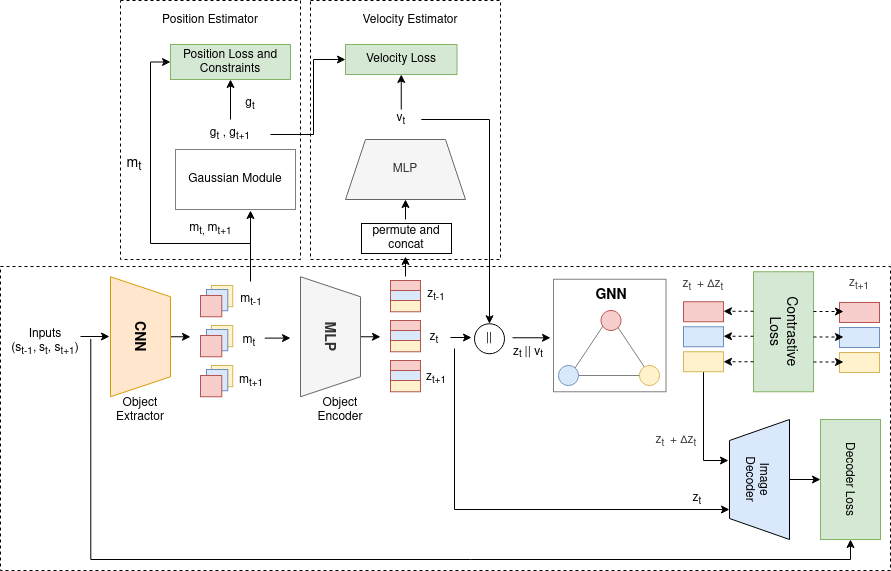}
    \caption{\label{Fig:icswm_model} Block diagram for learning an Interpretable Latent Space in Structured Models. Lower part of the figure is inspired from CSWM [Kipf. et al., 2020]}
    %~\cite{kipf&al20}} 
\end{figure*}

To achieve this, we introduce our model, \emph{Interpretable and Contrastive Learning of Structured World Models} (ICSWM) that learns a factored latent space transition model and estimates position and velocity. Our model is trained in an end-to-end unsupervised learning fashion such that we do not need any supervision on the positions and velocities of the objects.
% Similarly, $m_t= (m^k_t)^K_{k=1}$ represent the $k$ object masks for each of the $k$ objects in the environment.
% In addition to that we also want to learn position and velocity in order to add interpratability to the latent space.
Figure~\ref{Fig:icswm_model} shows the block diagram of our model. There are 5 basic blocks in the pipeline, 1) Object Extractor and Encoder, 2) Position Estimator, 3) Velocity Estimator, 4) Transition Model, and 5) Image Decoder. Correspondingly, we define various loss functions targeted to facilitate each module's learning. As our model is motivated from CSWM, we explicitly mention wherever we use or adapt any of the CSWM's modules.

\subsection{Object Extractor and Encoder}
The job of this module is to generate object masks and object encodings.
% We adapt the object extractor and encoder from CSWM.
An input state $s_t$ is processed using a CNN module, called the Object Extractor, that generates $K$ output channels each corresponding to the $K$ objects in our environment.
Next, we apply background subtraction to these channels by subtracting a learnable parameter $\theta$ shared among all channels. We call the resulting channels object masks denoted by $(m^k_t)^K_{k=1}$, where $K$ is the number of objects in the environment. In comparison to this CSWM does not do any background subtraction.

\begin{figure}[t]
    % \minipage{\linewidth}
      \centering
      \includegraphics[width=\linewidth]{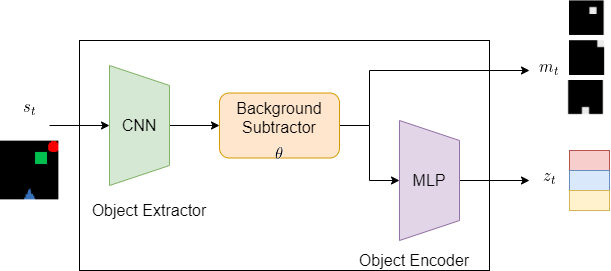}
    \caption{\label{Fig:block_diagram_Ob_Ex_En} Detailed diagram showing Object Extractor and Encoder}
\end{figure}

Similar to CSWM, we use an MLP to generate object encodings $(z^k_t)^K_{k=1}$ from object masks $(m^k_t)^K_{k=1}$. These encodings form our factorized latent space. The parameters of the object encoder are shared among all objects.
Figure~\ref{Fig:block_diagram_Ob_Ex_En} shows the detailed architecture and process of Object Extractor and Encoder.

\subsection{Position Constraints}\label{sec:model_pos}
In this section, we propose a novel method to estimate position of objects in an unsupervised fashion. Intuitively, we want to enforce three conditions, 1) we want each object mask to show a blob of pixels at the object's location (with everything else as black), 2) no two objects should overlap each other and 3) we want each object mask should contain only one such blob.
% Next, we use the object masks to get a numerical value corresponding to the positions of the objects.

\begin{figure}[tbp]
    % \minipage{\linewidth}
      \centering
      \includegraphics[width=\linewidth]{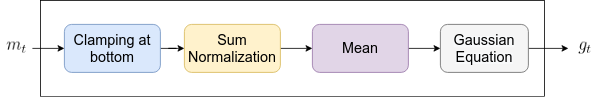}
    \caption{\label{Fig:block_diagram_Pos} Detailed diagram showing Position Estimator}
\end{figure}

For enforcing the first condition, we encourage each object mask to look like a 2D Gaussian distribution such that the mean of the distribution will give us the position of the object. Details of the position estimator are shown in Figure~\ref{Fig:block_diagram_Pos}. First, we clamp each object mask ($m^k_t$) by a small positive value ($10^{-8}$). Next, we normalize each mask independently by dividing it by $\sum_{i,j} m^k_t(i,j)$. For each object mask, we calculate the mean position of the pixels as $(\sum_i i* \sum_j m^k_t(i,j), \sum_j * \sum_i m^k_t(i,j))$. We use this mean position and a fixed variance $\sigma^2$ ($\sigma^2=0.6$ in our experiments) to create a Gaussian distribution denoted by $(g^k_t)^K_{k=1}$. Furthermore, the mean of $g^k_t$ gives the position of $k^{th}$ object and is denoted by $p^k_t$.
To ensure object masks look like Gaussian distributions, we add a position loss between each $m^k_t$ and $g^k_t$:
\begin{equation} \label{icswm:pos_gauss}
\begin{split}
 L_{p_{gauss}} = \frac{1}{K}\sum^{K}_{k=1} MSE(m^k_t, g^k_t).
\end{split}
\end{equation}

To enforce the second and third conditions, we add three constraints on the object masks. We add these constraints in the form of loss functions given in equation~\ref{icswm:pos_constraints}.
The first loss encourages each object mask to represent at least one object. 
The second loss coupled with the first one encourages each mask to detect at most one object. The third loss minimizes dot product between all object mask pairs to make sure object masks are mutually exclusive. Combining these three constraints encourages each objects mask to represent only one object.
\begin{equation} \label{icswm:pos_constraints}
\begin{split}
 L_{pos\_max} &= \frac{1}{K}\sum^{K}_{k=1} (1 - \max_{i,j} m^k_t(i,j)) \\
 L_{pos\_sum} &= \frac{1}{K}\sum^{K}_{k=1} |1 - \sum_{i,j} m^k_t(i,j)| \\
 L_{pos\_dot} &= \sum^{K}_{k=1} \sum^{K}_{l=k+1} m^k_t \cdot m^l_t
\end{split}
\end{equation}

The combined position loss becomes,
\begin{equation} \label{icswm:pos_final}
\begin{split}
 L_{pos} &= L_{pos\_gauss} + L_{pos\_sum} + L_{pos\_max} + L_{pos\_dot}
\end{split}
\end{equation}

\subsection{Velocity Constraint}
To make the model interpretable we want to estimate the velocity of the objects by adding simple domain knowledge. For this, we want to "tell" ICSWM+D the general relation between the position and velocity of an object. To this end, we train an MLP, called velocity estimator, that takes as input a permutation of both the current and previous state encodings,  i.e. $z_t$ and $z_{t-1}$, and estimates the velocity of objects.

Let us denote the concatenation of object encodings at time $t$ and $t-1$ by $z^k_{t,t-1} = [z^{k}_t || z^{k}_{t-1}]$. For predicting the velocity of the $k^{th}$ object, we give the permutation $[z^k_{t,t-1} || z^{k+1}_{t,t-1} || ~ \dots ~|| z^{K}_{t,t-1} || z^{1}_{t,t-1} || z^{2}_{t,t-1} ||~ \dots ~ || z^{k-1}_{t,t-1}]$ as input to the velocity MLP (for environments with no object interactions we provide only $[z^k_{t,t-1}]$ as input). 
To establish and enforce a relation between positions of object at time $t$ and $t+1$ and velocity of an object at time $t$, we encourage the predicted velocity to satisfy the constraint given by $p_{t+1} = p_t + v_t$. The corresponding loss function is given by,
\begin{equation} \label{icswm:vel_loss}
\begin{split}
 L_{vel} = \frac{1}{K}\sum^{K}_{k=1}  MSE(p^k_{t+1}, p^k_t + v^k_t)
\end{split}
\end{equation}
Note that the parameters of the Velocity MLP are shared among all objects. As CSWM does not predict the velocity so it does not have this module.

\subsection{Learning Transition Model using Graph Neural Network}
\label{subsec:cl}
Similar to CSWM, we use a Graph Neural Network (GNN)~\cite{Scarselli&al09} to learn the transition model.
The GNN predicts the translation vector $\Delta z_t$ representing change in the current latent space $z_t$.
The predicted latent representation of the next state is then given by $\hat{z}_{t+1}=z_t + \Delta z_t$.

We first create a fully connected graph with $K$ nodes representing the $K$ objects. Next, we use a concatenation of $z_t$ and $v_t$ as input features to the GNN (this is in contrast to CSWM that uses only $z_t$ as the input node features).
The transition model $T(z_t||v_t)$ is given as,
\begin{equation} \label{icswm:transition_function}
\begin{split}
 \Delta z_t &= T(z_t, v_t) = GNN((z^k_t || v^k_t)^K_{k=1}) \\
 \hat{z}_{t+1} &= (z^k_t +\Delta z^k_t)^K_{k=1}
\end{split}
\end{equation}
Here, $(\cdot||\cdot)$ represents the concatenation operation.
We first compute the edge embeddings of all edges using $f_{edge}$ and then compute node embeddings of all nodes using $f_{node}$. Since, we use a fully connected graph of objects, we do not do multiple iterations of the embedding computation as each node already has access to all other nodes.
We use the same node update function $f_{node}$ and an edge update function $f_{edge}$ as used in CSWM. $f_{node}$ and $f_{edge}$ are implemented using an MLP and message passing is given by,
\begin{equation} \label{icswm:gnn}
\begin{split}
 e^{(i,j)}_t &= f_{edge}(z^i_t, z^j_t) \\
 \Delta z^j_t &= f_{node}(z^j_t || v^j_t, \sum_{i, j} e^{(i,j)}_t)
\end{split}
\end{equation}
Next, we randomly sample a tuple of $(\tilde{s_t}, \tilde{s_{t+1}})$ from the experience buffer and get corresponding latent encoding as $(\tilde{z}_t, \tilde{z}_{t+1})$.
The encodings of the predicted next state and negative sample are then used in the contrastive loss given by,
\begin{equation} \label{icswm:loss}
\begin{split}
L_{con} &= d(z_t +  T(z_t, a_t), z_{t+1}) \\
 &~~~~ + max(0, \gamma - d(\tilde{z}_t, z_{t})) \\
 &~~~~ + max(0, \gamma - d(\tilde{z}_{t+1}, \hat{z}_{t+1}))
\end{split}
\end{equation}
Here $\gamma$ is a hyper parameter and we keep its value to be $1$ for all our experiments. 
There are three terms in the contrastive loss, the first term encourages $z_t$ to be close to the correct target and the last two terms encourage it to be different from the negative sample. We want to point out that our contrastive loss (Equation~\ref{icswm:loss}) adds a third term to the original loss of CSWM (Equation~\ref{cswm:loss}) to encourage the gradient of the negative term to pass through the GNN.  

\subsection{Image Decoder and pixel-based loss}
We define a two-phase curriculum of training. In the first phase, we focus on learning a good latent space using the position and contrastive loss.
Next, for phase 2 of the training we do two variations, 1) we add a velocity loss, and 2) we add velocity and a pixel-based loss. 
In our experiments variation 1 by ICSWM and variation 2 by ICSWM+D.
In both variations we also add position and contrastive loss in the phase 2 of the training process.
For our second variation, we use an image decoder that takes the latent space encoding $z_t$ as input and generates an image $\hat{s}_t$ that represents it. Our goal of adding an image decoder is to refine the latent space that has already been trained using position and contrastive loss. 
% It is important to note that the goal of image decoder is not to learn the transition model (like most of the work in MBRL) rather to convert latent space to image.
The decoder we use is the same as used in CSWM. The pixel-based loss calculates the regeneration error of the input images. In particular we calculate losses as $L_{dec}({s_t, \hat{s}_{t}})$ and $L_{dec}({s_{t+1}, \hat{s}_{t+1}})$, where

\begin{equation} \label{icswm:loss_d}
\begin{split}
L_{dec}(s,\hat{s}) &= BCE(s,\hat{s})
\end{split}
\end{equation}
Where BCE is the Binary Cross Entropy loss as used in CSWN~\cite{kipf&al20}.
The final loss for both phases of variation 2 is given by equation~\ref{icswm:loss_total} (for variation 1, we do not include $L_{dec}$ in $L_{phase_2}$). Here $\lambda_1$ and $\lambda_2$ are hyperparamters and in our experiments we keep $\lambda_1=10^{-1}$ and $\lambda_2=10^{-3}$.
\begin{equation} \label{icswm:loss_total}
\begin{split}
L_{phase_1} &= L_{pos} + L_{con} \\
L_{phase_2} &= L_{pos} + L_{con} + \lambda_1 L_{vel} + \lambda_2 L_{dec}
\end{split}
\end{equation}

%Template : https://www.overleaf.com/learn/latex/Algorithms#Algorithmic_package . Algorithm2e wala in this link

% \begin{algorithm}[H]
% \caption{PE\_CSWM}
% \SetAlgoLined
% \DontPrintSemicolon
% \textbf{Inputs:} $s_{t-1}$, $s_{t}$, $s_{t+1}$\newline
% $total\_epochs \gets 275$\;\newline
% $phase1\_epochs \gets 125$\;\newline
% $lambda1 \gets 0.1$\;\newline
% \for{$epoch = 1$}{$epoch <= total_epoch$}{$epoch++$}{$epoch += 1$\;}
% \end{algorithm}

% \begin{algorithm}
% \caption{PE-CSWM}
% \begin{algorithmic}
% % {$s_{t-1}, s_{t}, s_{t+1}$}
% $tot\_epochs \gets 275$
% $p1\_epochs \gets 125$
% \FOR{$epoch \leq tot\_epochs$}
%     \If{$condition = True$}
%         \State Do this
%         \If{$Condition \geq 1$}
%         \State Do that
%         \ElsIf{$Condition \neq 5$}
%         \State Do another
%         \State Do that as well
%         \Else
%         \State Do otherwise
%         \EndIf
%     \EndIf
%     \While{$something \not= 0$}  \Comment{put some comments here}
%         \State $var1 \leftarrow var2$  \Comment{another comment}
%         \State $var3 \leftarrow var4$
%     \ENDWHILE
% \ENDFOR
% \label{roy's loop}
    
% % \EndProcedure

% \end{algorithmic}
% \end{algorithm}

% https://shantoroy.com/latex/how-to-write-algorithm-in-latex/

% \input{model_old}
\section{Experiments}
We would like to answer the following questions in our experiments (a) Does the use of an interpretable latent space  help improve quality of predictions? (b) What is the impact of curriculum based training for incorporating decoder loss on the quality of predictions? (c) How accurately can our model predict the object positions and velocities? To answer these questions, we experimented with a variety of domains which include (a) Objects moving in a grid-world bouncing against the walls (b) bouncing objects which are also allowed to collide with each other resulting in object level interactions. We compare variations of ICSWM model, with variations of CSWM baseline. We describe details of our experiments next.

\subsection{Datasets}
We experiment with two different kinds of datasets. 

\begin{figure}[htbp]
    \minipage{0.5\linewidth}
      \centering
      \includegraphics[width=\linewidth]{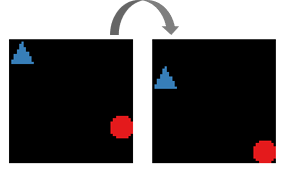}
      (a) 2-BS
    \endminipage\hfill
    \minipage{0.5\linewidth}
      \centering
      \includegraphics[width=\linewidth]{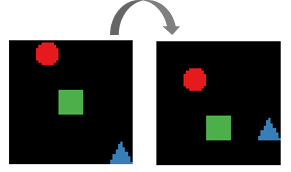}
      (b) 3-BS
    \endminipage
    \vspace{0.2cm}
    \minipage{0.5\linewidth}%
      \centering
      \includegraphics[width=\linewidth]{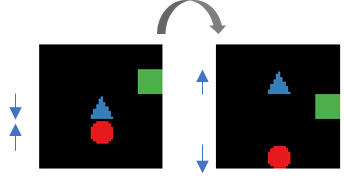}
      (c) 3-BSC
    \endminipage
    \minipage{0.4\linewidth}%
      \centering
      \includegraphics[width=\linewidth]{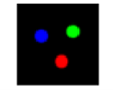}
      (d) 3-BP
    \endminipage
    \caption{\label{Test Environments} Sample observations of our test environments. a-b belong to the Non-Collision environments and c-d belong to Collision Environments. Arrows in c indicate the velocity directions of colliding objects}
    \label{fig:samples}
\end{figure}

\noindent
{\bf Bouncing Shapes:} Here, we model objects of different colors and shapes in a Grid environment, which can bounce along the wall. The objects do not collide with each other, so their motion is independent of other objects. Within this setting, we experiment with two different varations (a) five by five sized grid with two objects moving along columns. We refer to this as {\em 2-BouncingShapes (2-BS)} domain. (b) Five by five sized grid with three objects moving along columns. We refer to this as {\em 3-BouncingShapes (3-BS)} domain. To avoid collision, objects are placed in different columns in any given episode, and can only move along the column. Objects are initialized with with random position within the column, and start moving up or down with a constant (1-step) velocity unknown to the model. Once they hit the wall, they bounce back and start moving the same velocity but in the opposite direction. The goal is to be able to model the bouncing dynamics, and also predict the actual position and velocity (magnitude and sign) of every object in a future frame. See Figure~\ref{fig:samples} ((a),(b)) for example images from the two domains, respectively.

In the 2-BS domain, we created a dataset with (a) 1000 training episodes (b) 100 validation episodes (c) 100 test episodes, using a simulator. 2 Shapes were used : Red Circle, and Blue Triangle. In the 3-BS domain, we created a dataset with (a) 1000 training episodes (b) 100 validation episodes (c) 100 test episodes, using a simulator. The 3 Shapes were: Red Circle, Green cube, and Blue Triangle. All episodes for 2-BS and 3-BS have 100 steps each.

\noindent 
{\bf Bouncing Shapes with Collision:} Here, we model objects of different colors and shapes as before, but now they are allowed to collide with each other. In the first variation, referred to as {\em 3-Bouncing Shapes with Collision (3-BSC)} objects are placed in a five by five grid as before, and move along the columns. But now, one of the columns has more than one object in it, such that when they start moving, they will collide with each other. Two objects are said to collide with each other, if there next state position based on the current velocity happens to the be in the same grid cell. After collision, objects start moving in the opposite direction but with the same magnitude of velocity. Initial positions and velocity directions are determined randomly, for each object.  

Our final domain is {\em 3-body Physics (3-BP)} used earlier in the literature~\cite{kipf&al20}. The domain consists of 3 balls, moving in an environment, under the effect of gravity, and colliding with each other. The motion dynamics are not provided to the model. 
%For the purpose of our experiments, we changed the ball sizes (uniformly) so that they could fit our extractor size~\footnote{relaxing the size constraint is a direction for future work.}. Rest of the dynamics are the same as in the standard 3-body physics domain.
See Figure~\ref{fig:samples} ((c),(d)) for example images from 3-BSC, and 3-BP, respectively.

In the 3-BSC domain, we created a dataset with (a) 1000 training episodes (b) 100 validation episodes (c) 100 test episodes, using a simulator (All episodes have 100 steps each). The 3 Shapes were: Red Circle, Green cube, and Blue Triangle. For 3-BP, we created a dataset with (a) 5000 training episodes (b) 100 validation episodes (c) 1000 test episodes, using a simulator provided by ~\cite{kipf&al20} (All episodes have 10 steps each).

%Environment Samples

\subsection{Models Compared}
We experiment with the following variations of our model: \\

\noindent
{\bf ICSWM Variations:} 
\begin{itemize}
\item {\em ICSWM:} This is our model with an Interpretable Contrastive Loss in the latent Space as describe in Section~\ref{sec:model}. 
\item {\em ICSWM+D:} This refers to the ICSWM model with an additional decoder/pixel based loss imposed using curriculum training. 
\item  {\em ICSWM+D-C:} This refers to the ICSWM model with an additional decoder/pixel based loss imposed using curriculum training, but without any contrastive loss in the latent space. 
\end{itemize}
In all our models, we use the additional negative term in the contrastive loss. We next explain the variations of the Baseline model. \\

\noindent 
{\bf CSWM (Baseline) Variations:} 
\begin{itemize}
\item {\em  CSWM:} This is the contrastive loss based model as proposed by Kipf et al.~\cite{kipf&al20}.
\item {\em CSWM+N:} This refers to the CSWM model with an additional negative term in contrastive loss as explained in Section~\ref{subsec:cl}. 
\item {\em CSWM+D:} This refers to the CSWM model with an additional decoder/pixel based loss imposed using curriculum training as explained in 
Section~\ref{sec:model}. 
\item {\em CSWM+N+D:} This is the combination of CSWM+N and CSWM+D models, with both additional negative loss term, as well as, decoder based loss incorporated during training.
\end{itemize}

It is important to note that we tried variations of the CSWM model, such as adding Decoder based loss in the pixel space, or having additional negative loss term, to examine if the gains in the ICSWM model are because of these small enhancements, or primarily due to our interpretable latent space model. This was important to separate out the impact of various architectural changes proposed in this paper.
\subsection{Evaluation Metrics}
We use the following evaluation metrics.
\begin{itemize}
    \item {\bf Hits@k}: This is a ranking-based metric which evaluates the performance of the model directly in the latent space. Hits @ k measures what fraction of times the predicted latent state, i.e.,  $\hat{z}_{t+1}$, is ranked top amongst the $k$-ranked latent state representations, where the reference states are chosen randomly from the experience buffer, and the ranking is based on inverse distance from the latent space representation of $s_{t+1}$ i.e. $z_{t+1}$.
    \item {\bf MRR (Mean Reciprocal Rank)}
    This is another ranking-based metric that operates directly on latent state representations. MRR is the mean of the reciprocal of rank of the predicted next state encoded representation, $\hat{z}_{t+1}$, where the reference states are the latent space representations of randomly chosen states from experience buffer, and the ranking is computed using  inverse distance from the latent space representation of $s_{t+1}$ i.e. $z_{t+1}$, as in Hits@k.

    %Mathematically, MRR is given by: 
%\[MRR =  \frac{1}{N} \sum_{n=1}^{N} \frac{1}{rank_{n}} \]
    
    % \item {\bf BCE (Binary Cross-Entropy):} This represents the binary cross entropy loss between the $\hat{s}_{t+1}$ and $s_{t+1}$ measured in the decoded pixel space. This is useful to measure the actually quality of reconstruction, after the predicted latent representation is passed through a decoder. \rgcom{May remove this since we never showed any graph for this}
\end{itemize}
\subsection{Training Methodology}
We trained all the models for 275 epochs, and picked up the best using a validation set. For models which exploit a two-phase training using decoder loss, we trained the first phase for a set of 125 epochs, and then trained for second phase for maximum of 150 epochs. For CSWM and ICSWM, we trained the decoder after 275 epochs, freezing the rest of the pipeline for another 150 epochs to report the reconstructions as in figure~\ref{fig:3-BS_recon}. The other hyper-parameters such as learning algorithm and the learning rate, were kept the same as in the CSWM paper~\cite{kipf&al20}. All our experiments were run on a high performance facility having NVIDIA K-40 GPU. Training time of models varied between 3 to 9 hours. The reported numbers are averaged over 3 different training runs.

\subsection{Quantitative Results}
\subsubsection{Baseline Models}
\begin{figure}
\centering
\includegraphics[scale = 0.425]{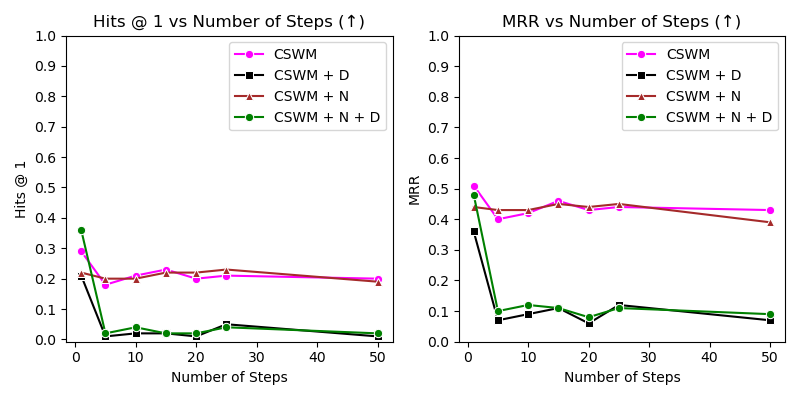}
\caption{On 2-BouncingShapes. Comparison between performance of variants of CSWM on Hits@1 and MRR. Higher is better.}
\label{fig:base_cmp_2-BS}
\end{figure}

\begin{figure}
\centering
\includegraphics[scale = 0.425]{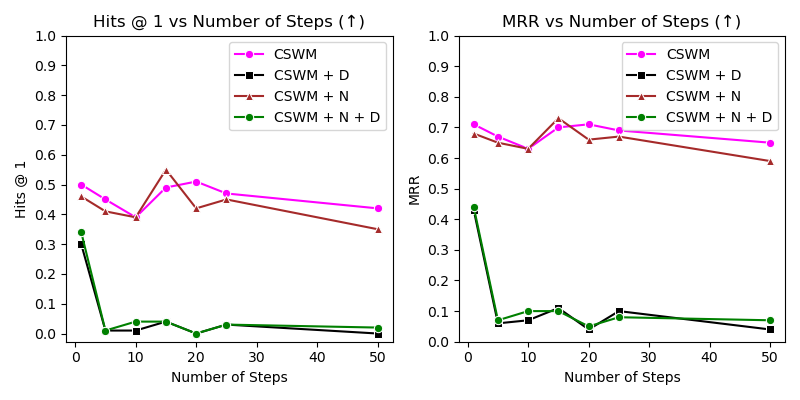}
\caption{On 3-BouncingShapes. Comparison between performance of variants of CSWM on Hits@1 and MRR. Higher is better.}
\label{fig:base_cmp_3-BS}
\end{figure}

We first wanted to compare which variants of the baseline are most robust for further comparison with our ICSWM model, and its variations. To examine this, we compared the 4 baseline variations on the BouncingShapes domains, i.e., 2-BS and 3-BS. Results are presented in Figures~\ref{fig:base_cmp_2-BS}, and ~\ref{fig:base_cmp_3-BS}, respectively. We measure both Hits@1 and MRR, as number of prediction steps is varied from 1 to 50. Clearly, we can see the models using an additional decoder loss, i.e., CSWM+D and CSWM+N+D, perform significantly worse compared to non-decoder based models for both Hits@1 and MRR, for prediction at all future steps. This means that CSWM model is not able to exploit the additional decoder based curriculum training to its advantage. Further, performance of CSWM and CSWM+N, is comparable to each other in both the experiments. We therefore, decided to only use CSWM as our primary baseline for our future set of experiments. 

\noindent
%%Baseline establishment
\subsubsection{Bouncing Shapes}  
\begin{figure}
\begin{center}
\includegraphics[scale = 0.425]{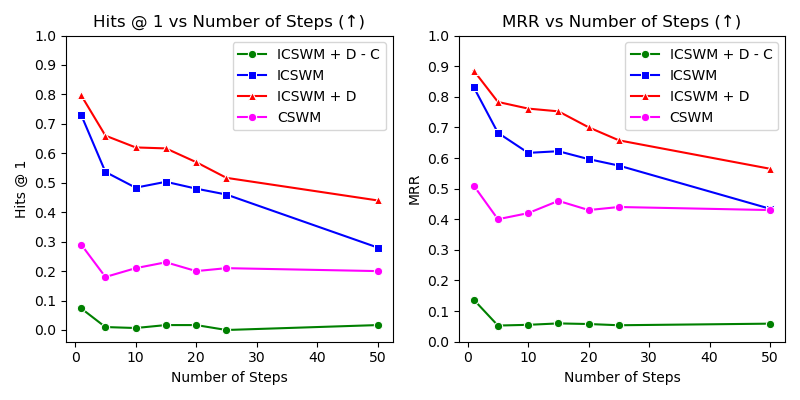}
\end{center}
\caption{Comparison between ICSWM variants and CSWM baseline using Hits @ 1 and MRR. For both metrics, higher is better. Domain: 2-BS}
\label{fig:2-BS}
\end{figure}

\begin{figure}
\begin{center}
\includegraphics[scale = 0.425]{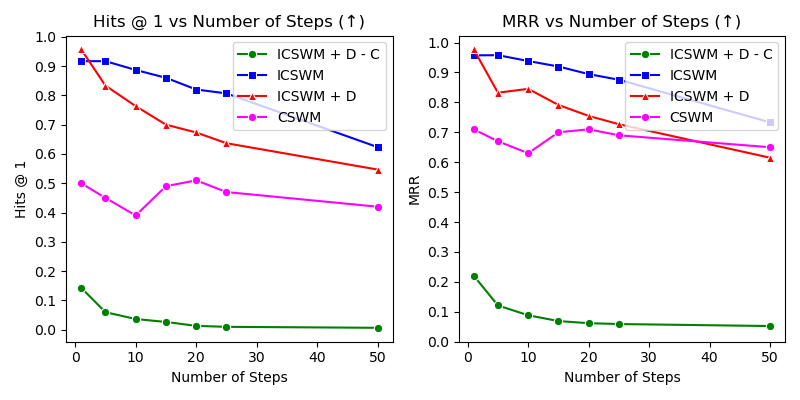}
\end{center}
\caption{Comparison between ICSWM variants and CSWM baseline using Hits @ 1 and MRR. For both metrics, higher is better. Domain: 3-BS}
\label{fig:3-BS}
\end{figure}

%Comparison plots
Figures~\ref{fig:2-BS},~\ref{fig:3-BS} compare the performance of ICSWM variants with CSWM baseline, for 2-BS and 3-BS domains, respectively. We compare on both Hits@1 and MRR, as the number of future steps is varied from 1 to 50. We see that both ICSWM and ICSWM+D perform better than CSWM baseline, at all steps~\footnote{ICSWM+D performs marginally worse than CSWM on 3-BS in MRR for last few steps but this may be simply due to variance} , for both the metrics, in both the domains. This clearly points to the power of having an interpretable latent space model. The gain is significantly more at initial prediction steps, and the difference decreases as we increase the number of steps. We hypothesise that as the number of steps is increased to 50, all the models gradually approach random behavior, which is also confirmed by the flattening nature of the curves towards the end. The comparison between ICSWM and ICSWM+D is interesting. Where the former does better on 2-BS, the latter does somewhat better on 3-BS. 
%Investigating further into the reasons behind this relative difference in performance is a direction for future work. 
ICSWM+D-C performs the worst which points to the importance of having a contrastive loss.

\subsubsection{Bouncing Shapes with Collision}

\begin{figure}
\begin{center}
\includegraphics[scale = 0.425]{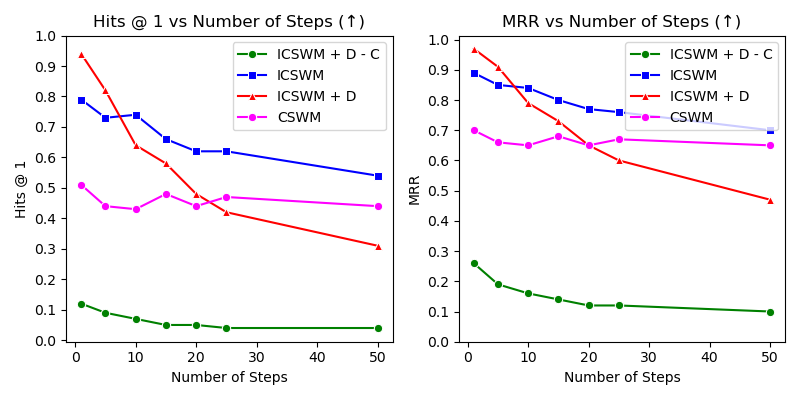}
\end{center}
\caption{Comparison between ICSWM variants and CSWM baseline using Hits @ 1 and MRR. For both metrics, higher is better. Domain: 3-BSC}
\label{fig:3-BSC}
\end{figure}

Figure~\ref{fig:3-BSC} compares the performance of ICSWM variants with CSWM baseline for 3-BSC domain. We compare on both Hits@1 and MRR, as the number of future steps is varied from 1 to 50. On 2-BSC, ICSWM beats CSWM baseline at all time steps for both the metrics. Interestingly, while ICSWM+D does best for first few steps, its performance degrades quickly, and it becomes worse than both ICSWM and CSWM by the end of 50 steps. Understanding this behaviour of our model, when used with decoder loss, is a direction for future investigation. ICSWM+D-C performs worse than all since it is not able to exploit the contrastive loss.

\begin{figure}
\begin{center}
\includegraphics[scale = 0.425]{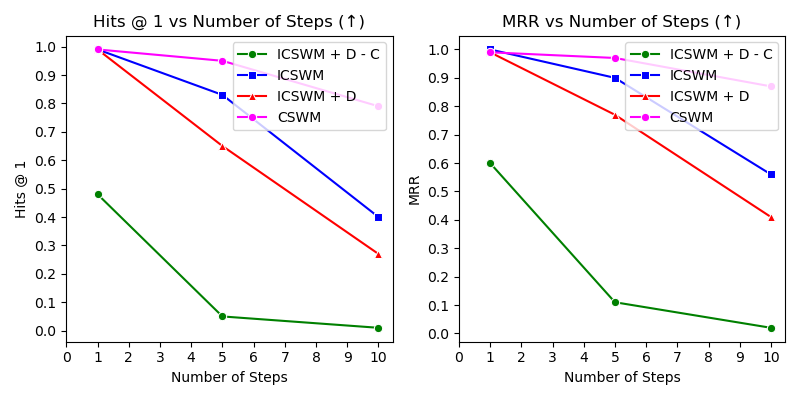}
\end{center}
\caption{Comparison between ICSWM variants and CSWM baseline using Hits @ 1 and MRR. For both metrics, higher is better. Domain: 3-BP}
\label{fig:3-BP}
\end{figure}

Figure~\ref{fig:3-BP} compares the performance of ICSWM variants with CSWM baseline for 3-BP domain, on both Hits@1 and MRR. Here, CSWM is the best performing model, followed by ICSWM and ICSWM+D respectively. While all the models perform equally well for 1-step prediction, our models start performing worse (compared to baseline) for future steps. This was quite intriguing result given the superior performance of ICSWM variants on other 3 domains. Further investigation revealed that, while 2-BS, 3-BS and 3-BSC domains are heavily dependent on localization of objects with respect to the image frame, the dynamics in 3-BP are purely governed by inter-object interactions due to forces of gravitational pull. While our models are good at localizing due to interpretable position and velocity constraints, the results in 3-BP points that though effective, they may not be as good as CSWM in terms of capturing inter-object interactions. We believe this may be due to a relatively simple velocity module in our case, which is key to capturing object interactions; GNN simply acts on the position based feature maps, and the velocity output by the velocity module in our models. Any error in velocity prediction step is going to make GNN less effective in ICSWM models. 

On the other hand, CSWM fails quite badly when predicting motion dynamics requires localization of objects in the frame, as demonstrated by results on 2-BS, 3-BS and 3-BSC. This is also evident in our qualitative results, where we show that our object feature maps are much more interpretable and can localize the objects fairly well, whereas CSWM fails to do so, except for some very simple domains. Capturing localization behaviour may be important in Atari games, such as pong, where our model is expected to do well on this aspect~\footnote{Performing these experiments is a part of future work.}  Improving the object interaction component of our model, while still keeping intact its localization capability is a direction for future work.

\subsubsection{Qualitative Results}

\begin{figure}[htbp]
    \begin{center}
    \minipage{0.68\linewidth}
      \centering
      \includegraphics[width=\linewidth]{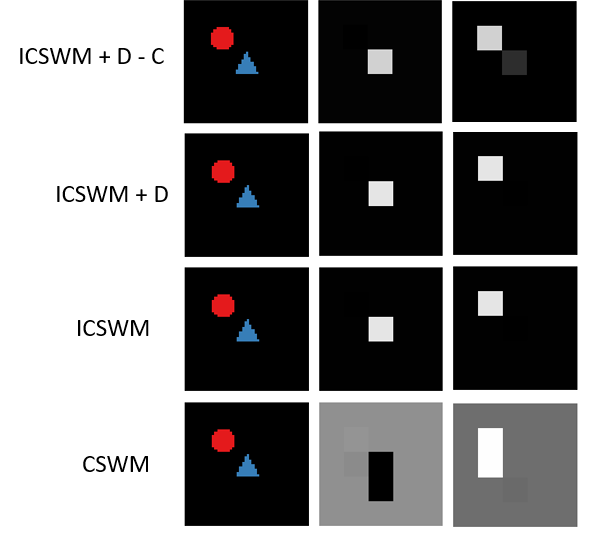}
      \caption{(a) Object feature maps : 2-BS}
    \endminipage
    \hspace{0.7cm}
    \vspace{0.2cm}
    
    \minipage{0.80\linewidth}
      \centering
      \includegraphics[width=\linewidth]{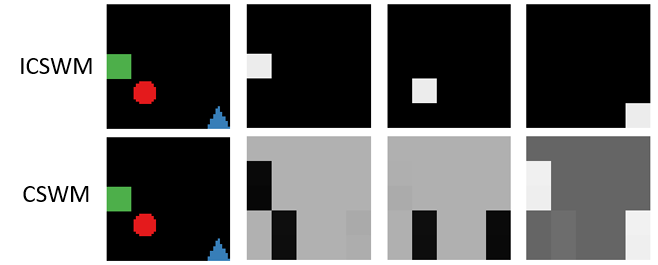}
      \caption{(b) Object feature maps: 3-BS}
      \vspace{0.2cm}
    %\endminipage
    %\end{center}
%\end{figure}
%\begin{figure}    
    %\minipage{0.5\linewidth}%
      \centering
      \includegraphics[width=\linewidth]{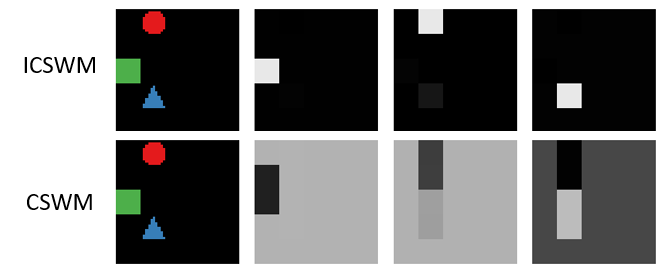}
      \caption{(c) Object feature maps: 3-BSC} 
      \vspace{0.2cm}
    %\endminipage\hspace{0.5cm}
    %\minipage{0.5\linewidth}%
      \centering
      \includegraphics[width=\linewidth]{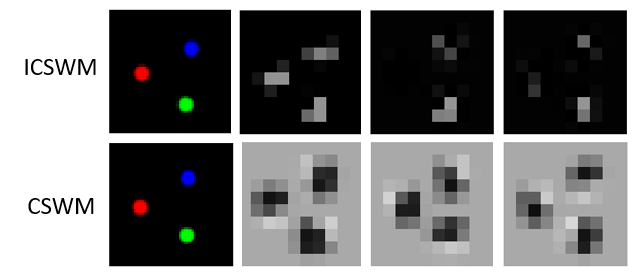}
      \caption{(d) Object feature maps: 3-BP}
    \endminipage
    \end{center}
    \caption{ Comparison of feature maps on the 4 environments, viz. 2-BS, 3-BS, 3-BSC and 3-BP. For fig. 10 (a), the first row is the feature maps of ICSWM + D - C, second row shows feature maps of ICSWM + D, third shows ICSWM and the last row belongs to CSWM. For fig 10 (b) - (d), only ICSWM (first row) and CSWM (second row) have been compared since they are the best-performing versions of ours and CSWM}
    \label{fig:features}
\end{figure}

Figure~\ref{fig:features} shows the feature maps constructed by ICSWM variants and the CWSM for all the domains. The feature maps constructed by ICSWM and its variants are much sharper, and can separate the object much better, compared to CSWM. Interestingly, this is also true for 3-BP where the performance of CSWM is better than ICSWM variants. None of the algorithms can achieve perfect feature maps in this case, but ours are much closer to what they should be in a disentangled representation compared to the CSWM baseline. 

\begin{figure}
\begin{center}
\includegraphics[scale = 0.70]{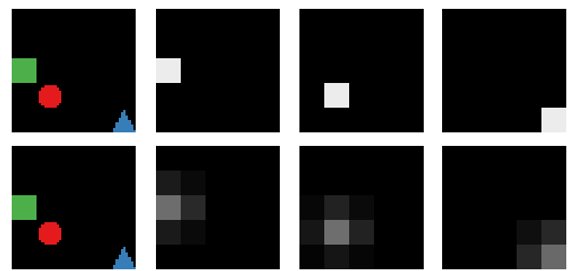}
\end{center}
\caption{Sample 2D Gaussian distribution (discretized). The two rows represent the object feature maps, and Gaussian distributions, respectively. First column represents the true image. Next 3 columns represent 3 feature maps (for 3 objects). Domain: 3-BS}
\label{fig:3-BS_gaussian}
\end{figure}

\begin{figure}
\begin{center}
\includegraphics[scale = 0.675]{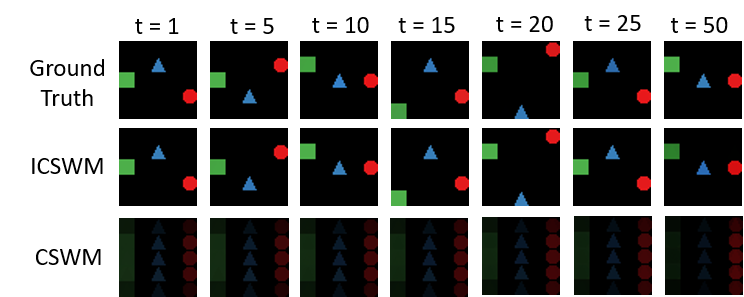}
\end{center}
\caption{Pixel Space Reconstructions for an episode shown at time steps 1, 5, 10, 15, 20, 25 and 50 (left to right order) given only the observation at time 0. Domain: 3-BS}
\label{fig:3-BS_recon}
\end{figure}

Figure~\ref{fig:3-BS_gaussian} depicts the comparison with 2D Gaussian distribution for the 3-BS domain for ICSWM. Clearly, there is a close resemblance between the Gaussian distribution and the actual object feature map detected by the algorithm. Finally, for illustration, we also trained a separate decoder network for ICSWM and CSWM, for maximum 150 epochs, after freezing the parameters of the latent model. Reconstructions were made using the decoder trained in this manner for 3-BS domain. Looking at the Figure~\ref{fig:3-BS_recon}, it is clear that ICSWM can do a much finer task of re-construction even at 50 steps, compared to CSWM baseline, which is quite ineffective in this task.% We also conducted a small experiment to determine how correctly our gaussian map can spit position. %We find that our gaussian map can spit position with an average standard error of $0.83 \pm 0.40$.

\section{Conclusion and Future Work}
In this paper we presented a novel approach to learn an interpretable latent space in structured models for video prediction. Our model builds on earlier work by Kipf et al. which uses a contrastive loss to train the model. We introduce novel position and velocity based constraints, inspired from generic physical laws, to enhance their latent space model. 
%learning loss in the latent space. 
%Our model is capable of estimating position and velocities of objects. We argued the efficacy of our approach over the baseline by showing our results on 3 environments.
As an important step in this direction, our model can learn highly localized feature maps (compared to the base model), but faces some difficulty in capturing complex interactions (like in 3-body physics). We outperform the baseline in 3 out of 4 domains that we experiment with in terms of latent space predictions. Our learned feature maps are highly interpretable and closely resemble actual object positions.

In terms of future work, one line of research is to develop a model that combines the strengths of both the models. Another future direction is to enhance our model to learn position and velocities in environments with objects of varying sizes (like that in OpenAI-Atari Pong and Breakout).

\section*{Acknowledgements}
We thank IIT Delhi HPC facility
%\footnote{\emph{http://supercomputing.iitd.ac.in}} 
for computational resources. Parag Singla and Guy Van den Broeck are supported by the DARPA Explainable Artificial Intelligence (XAI) Program \#N66001-17-2-4032. Parag Singla is supported by IBM SUR awards, and Visvesvaraya Young Faculty Fellowship by Govt. of India. Guy Van den Broeck is supported by NSF grants \#IIS-1943641, \#IIS-1633857, \#CCF-1837129, a Sloan Fellowship, and gifts by Intel and Facebook Research. Vishal Sharma is supported by TCS Research Scholar Fellowship. Any opinions, findings, conclusions or recommendations expressed in this paper are those of the authors and do not necessarily reflect the views or official policies, either expressed or implied, of the funding agencies.

%\newpage
\bibliographystyle{named}
\bibliography{main}

\begin{thebibliography}{}

\bibitem[\protect\citeauthoryear{Chen \bgroup \em et al.\egroup
  }{2017}]{cheng&al17}
Xiongtao Chen, Wenmin Wang, Jinzhuo Wang, and Weimian Li.
\newblock Learning object-centric transformation for video prediction.
\newblock In {\em Proceedings of the 25th ACM International Conference on
  Multimedia}, MM '17, page 1503–1512, New York, NY, USA, 2017. Association
  for Computing Machinery.

\bibitem[\protect\citeauthoryear{Chiappa \bgroup \em et al.\egroup
  }{2017}]{chiappaetal17}
Silvia Chiappa, S{\'e}bastien Racaniere, Daan Wierstra, and Shakir Mohamed.
\newblock Recurrent environment simulators.
\newblock In {\em In Proceedings of International Conference on Learning
  Representations (ICLR)}, 2017.

\bibitem[\protect\citeauthoryear{Denton and Birodkar}{2017}]{Denton&al17}
Emily~L Denton and vighnesh Birodkar.
\newblock Unsupervised learning of disentangled representations from video.
\newblock In I.~Guyon, U.~V. Luxburg, S.~Bengio, H.~Wallach, R.~Fergus,
  S.~Vishwanathan, and R.~Garnett, editors, {\em Advances in Neural Information
  Processing Systems}, volume~30. Curran Associates, Inc., 2017.

\bibitem[\protect\citeauthoryear{Ha and Schmidhuber}{2018}]{haetal18}
David Ha and J{\"u}rgen Schmidhuber.
\newblock World models.
\newblock {\em arXiv preprint arXiv:1803.10122}, 2018.

\bibitem[\protect\citeauthoryear{Henaff \bgroup \em et al.\egroup
  }{2017}]{henaffetal17}
Mikael Henaff, William~F Whitney, and Yann LeCun.
\newblock Model-based planning with discrete and continuous actions.
\newblock {\em arXiv preprint arXiv:1705.07177}, 2017.

\bibitem[\protect\citeauthoryear{Hou \bgroup \em et al.\egroup
  }{2019}]{Hou&al19}
Ruibing Hou, Hong Chang, Bingpeng Ma, and Xilin Chen.
\newblock Video prediction with bidirectional constraint network.
\newblock In {\em 2019 14th IEEE International Conference on Automatic Face
  Gesture Recognition (FG 2019)}, pages 1--8, 2019.

\bibitem[\protect\citeauthoryear{Hsieh \bgroup \em et al.\egroup
  }{2018}]{Hsieh&al18}
Jun-Ting Hsieh, Bingbin Liu, De-An Huang, Li~Fei-Fei, and Juan~Carlos Niebles.
\newblock Learning to decompose and disentangle representations for video
  prediction.
\newblock In {\em Proceedings of the 32nd International Conference on Neural
  Information Processing Systems}, NIPS'18, page 515–524, Red Hook, NY, USA,
  2018. Curran Associates Inc.

\bibitem[\protect\citeauthoryear{Jaques \bgroup \em et al.\egroup
  }{2020}]{jaques&al20}
Miguel Jaques, Michael Burke, and Timothy~M. Hospedales.
\newblock Physics-as-inverse-graphics: Joint unsupervised learning of objects
  and physics from video.
\newblock In {\em In Proceedings of International Conference on Learning
  Representations (ICLR)}, 2020.

\bibitem[\protect\citeauthoryear{Jin \bgroup \em et al.\egroup
  }{2020}]{Jin&al20}
Beibei Jin, Yu~Hu, Qiankun Tang, Jingyu Niu, Zhiping Shi, Yinhe Han, and
  Xiaowei Li.
\newblock Exploring spatial-temporal multi-frequency analysis for high-fidelity
  and temporal-consistency video prediction.
\newblock {\em CoRR}, abs/2002.09905, 2020.

\bibitem[\protect\citeauthoryear{Kaiser \bgroup \em et al.\egroup
  }{2020}]{kaiser&al20}
Lukasz Kaiser, Mohammad Babaeizadeh, Piotr Milos, Blazej Osinski, Roy~H.
  Campbell, Konrad Czechowski, Dumitru Erhan, Chelsea Finn, Piotr Kozakowski,
  Sergey Levine, Ryan Sepassi, George Tucker, and Henryk Michalewski.
\newblock Model-based reinforcement learning for atari.
\newblock In {\em In Proceedings of International Conference on Learning
  Representations (ICLR)}, 2020.

\bibitem[\protect\citeauthoryear{Kipf \bgroup \em et al.\egroup
  }{2020}]{kipf&al20}
Thomas Kipf, Elise van~der Pol, and Max Welling.
\newblock Contrastive learning of structured world models.
\newblock In {\em In Proceedings of International Conference on Learning
  Representations (ICLR)}, 2020.

\bibitem[\protect\citeauthoryear{Kosiorek \bgroup \em et al.\egroup
  }{2018}]{Kosiorek&al18}
Adam Kosiorek, Hyunjik Kim, Yee~Whye Teh, and Ingmar Posner.
\newblock Sequential attend, infer, repeat: Generative modelling of moving
  objects.
\newblock In S.~Bengio, H.~Wallach, H.~Larochelle, K.~Grauman, N.~Cesa-Bianchi,
  and R.~Garnett, editors, {\em Advances in Neural Information Processing
  Systems}, volume~31. Curran Associates, Inc., 2018.

\bibitem[\protect\citeauthoryear{Oh \bgroup \em et al.\egroup
  }{2015}]{ohetal15}
Junhyuk Oh, Xiaoxiao Guo, Honglak Lee, Richard Lewis, and Satinder Singh.
\newblock Action-conditional video prediction using deep networks in atari
  games.
\newblock {\em Proceedings of the 28th International Conference on Neural
  Information Processing Systems}, pages 2863--2871, 2015.

\bibitem[\protect\citeauthoryear{Oprea \bgroup \em et al.\egroup
  }{2020}]{Oprea&all20}
Sergiu Oprea, Pablo Martinez{-}Gonzalez, Alberto Garcia{-}Garcia,
  John~Alejandro Castro{-}Vargas, Sergio Orts{-}Escolano,
  Jos{\'{e}}~Garc{\'{\i}}a Rodr{\'{\i}}guez, and Antonis~A. Argyros.
\newblock A review on deep learning techniques for video prediction.
\newblock {\em CoRR}, abs/2004.05214, 2020.

\bibitem[\protect\citeauthoryear{Scarselli \bgroup \em et al.\egroup
  }{2009}]{Scarselli&al09}
Franco Scarselli, Marco Gori, Ah~Chung Tsoi, Markus Hagenbuchner, and Gabriele
  Monfardini.
\newblock The graph neural network model.
\newblock {\em IEEE Transactions on Neural Networks}, 20(1):61--80, 2009.

\bibitem[\protect\citeauthoryear{Shouno}{2020}]{Shouno20}
Osamu Shouno.
\newblock Photo-realistic video prediction on natural videos of largely
  changing frames.
\newblock {\em CoRR}, abs/2003.08635, 2020.

\bibitem[\protect\citeauthoryear{Srivastava \bgroup \em et al.\egroup
  }{2015}]{Srivastava&al15}
Nitish Srivastava, Elman Mansimov, and Ruslan Salakhutdinov.
\newblock Unsupervised learning of video representations using lstms.
\newblock In {\em In Proceedings of International Conference on Machine
  Learning (ICML)}, 2015.

\bibitem[\protect\citeauthoryear{Sun \bgroup \em et al.\egroup
  }{2018}]{sunetal18}
Chen Sun, Abhinav Shrivastava, Carl Vondrick, Kevin Murphy, Rahul Sukthankar,
  and Cordelia Schmid.
\newblock Actor-centric relation network.
\newblock In {\em Proceedings of the European Conference on Computer Vision
  (ECCV)}, pages 318--334, 2018.

\end{thebibliography}
\newpage
\newpage
\clearpage
\appendix
\section{Appendix}
\subsection{Architecture}
\begin{enumerate}
    \item \textbf{Object Extractor and Encoder:} 
            \begin{itemize}
                \item \textbf{Object Extractor:} Object Extractor is a CNN based network that takes input as a single RGB image of size 50 x 50 x 3 and returns $K$ feature maps. There are two types of extractors viz. small and medium. In small extractor, input RGB image is passed through a convolutional layer with kernel size 10, stride 10 and 64 filters, followed by a 2D batch norm layer, relu activation function, another convolutional layer with kernel size 1, stride 1 and $K$ filters where $K$ is the number of object feature maps and sigmoid activation function. Medium extractor differs only in convolutional layer specifications, however the overall flow remains same. First convolutional layer of medium extractor has a kernel size 9, padding of 4 and 64 filters and the second convolutional layer has kernel size 5, stride 5 and $K$ filters where $K$ is the number of object feature maps. For our experiments on 2-BS, 3-BS and 3-BSC, we use small eactor. However, for 3-BP we use medium extractor.
                \item \textbf{Background Subtractor:} It is a single parameter that is subtracted from the outputs of the object extractor. Note that the parameter is tied across all the output channels of object extractor.
                \item \textbf{Object Encoder:} It is a MLP network that takes input as flattened feature maps, one at a time and converts it into its latent space representation. The flattened feature map is passed through fully connected layer 1 followed by relu activation, fully connected layer 2, layer norm, relu activation, fully connected layer 3 with 2 units. We use the same embedding dimension of 2 for all experiments. 
            \end{itemize}
    \item \textbf{Position Estimator:} Position extractor is a non-neural architecture and details can be found in section~\ref{sec:model}.
    \item \textbf{Velocity Estimator} We use a three layer MLP with 4 units in layer 1 followed by relu activation function, 2 units in layer 2, relu activation and 1 unit in final layer of velocity module. This setting is used for experiments on 2-BS and 3-BS. For 3-BSC and 3-BP, we use a three layer MLP with 8 units in layer 1 followed by relu activation function, 4 units in layer 2, relu activation and 1 unit in final layer of velocity module for 3-BSC and 2 unit in final layer for 3-BP.
    \item \textbf{Transition Model} We use GNN for capturing interactions among objects. The message passing formulation is given in section~\ref{sec:model}. The architecture of MLP for $f_{node}$ and $f_{edge}$ is same as that in CSWM ~\cite{kipf&al20}. The only change we do is to substitute the action in the case of CSWM with our velocity as predicted by the velocity module.
    \item \textbf{Image Decoder} We use same decoder as used in CSWM~\cite{kipf&al20}.

\clearpage
\begin{table*}[!htp]\centering
\caption{Mean Value}\label{tab: }
\scriptsize
\begin{tabular}{lrrrrrrrrr}\toprule
& & & & & & & & \\
{\textbf{2-BS}} & & & & & & & & \\
\textbf{Num Steps}
% \multirow{\textbf{Num Steps}}
 &\multicolumn{2}{c}{\textbf{ICSWM+D-C}} &\multicolumn{2}{c}{\textbf{ICSWM}} &\multicolumn{2}{c}{\textbf{ICSWM + D}} &\multicolumn{2}{c}{\textbf{CSWM}} \\\cmidrule{2-9}
& \textbf{Hits @ 1} &\textbf{MRR} &\textbf{Hits@1} &\textbf{MRR} &\textbf{Hits @ 1} &\textbf{MRR} &\textbf{Hits @ 1} &\textbf{MRR} \\ \cmidrule{1-9}
1 &0.07 &0.14 &0.73 &0.83 &0.8 &0.89 &0.29 &0.51 \\\cmidrule{1-9}
5 &0.01 &0.05 &0.54 &0.68 &0.66 &0.78 &0.18 &0.4 \\\cmidrule{1-9}
10 &0.01 &0.06 &0.48 &0.62 &0.62 &0.76 &0.21 &0.42 \\\cmidrule{1-9}
15 &0.02 &0.06 &0.5 &0.62 &0.62 &0.75 &0.23 &0.46 \\\cmidrule{1-9}
20 &0.02 &0.06 &0.48 &0.6 &0.57 &0.7 &0.2 &0.43 \\\cmidrule{1-9}
25 &0 &0.05 &0.46 &0.58 &0.52 &0.66 &0.21 &0.44 \\\cmidrule{1-9}
50 &0.02 &0.06 &0.28 &0.43 &0.44 &0.56 &0.2 &0.43 \\\cmidrule{1-9}
& & & & & & & & \\
{\textbf{3-BS}} & & & & & & & & \\
\textbf{Num\_Steps}
% \multirow{\textbf{Num\_Steps}}
&\multicolumn{2}{c}{\textbf{ICSWM +D - C}} &\multicolumn{2}{c}{\textbf{ICSWM}} &\multicolumn{2}{c}{\textbf{ICSWM + D}} &\multicolumn{2}{c}{\textbf{CSWM}} \\\cmidrule{2-9}
&\textbf{Hits @ 1} &\textbf{MRR} &\textbf{Hits@1} &\textbf{MRR} &\textbf{Hits @ 1} &\textbf{MRR} &\textbf{Hits @ 1} &\textbf{MRR} \\\cmidrule{1-9}
1 &0.21 &0.26 &0.1 &0.05 &0.03 &0.02 &0.5 &0.71 \\\cmidrule{1-9}
5 &0.09 &0.12 &0.09 &0.04 &0.14 &0.08 &0.45 &0.67 \\\cmidrule{1-9}
10 &0.05 &0.06 &0.09 &0.05 &0.22 &0.14 &0.39 &0.63 \\\cmidrule{1-9}
15 &0.03 &0.03 &0.09 &0.05 &0.26 &0.19 &0.49 &0.7 \\\cmidrule{1-9}
20 &0.01 &0.02 &0.07 &0.04 &0.27 &0.21 &0.51 &0.71 \\\cmidrule{1-9}
25 &0 &0.01 &0.05 &0.05 &0.32 &0.25 &0.47 &0.69 \\\cmidrule{1-9}
50 &0.01 &0 &0.16 &0.16 &0.4 &0.32 &0.42 &0.65 \\\cmidrule{1-9}
& & & & & & & & \\
{\textbf{3-BSC}} & & & & & & & & \\
\textbf{Num\_Steps}
% \multirow{\textbf{Num\_Steps}}
&\multicolumn{2}{c}{\textbf{ICSWM +D - C}} &\multicolumn{2}{c}{\textbf{ICSWM}} &\multicolumn{2}{c}{\textbf{ICSWM + D}} &\multicolumn{2}{c}{\textbf{CSWM}} \\\cmidrule{2-9}
&\textbf{Hits @ 1} &\textbf{MRR} &\textbf{Hits@1} &\textbf{MRR} &\textbf{Hits @ 1} &\textbf{MRR} &\textbf{Hits @ 1} &\textbf{MRR} \\ \cmidrule{1-9}
1 &0.12 &0.26 &0.79 &0.89 &0.94 &0.97 &0.51 &0.7 \\\cmidrule{1-9}
5 &0.09 &0.19 &0.73 &0.85 &0.82 &0.91 &0.44 &0.66 \\\cmidrule{1-9}
10 &0.07 &0.16 &0.74 &0.84 &0.64 &0.79 &0.43 &0.65 \\\cmidrule{1-9}
15 &0.05 &0.14 &0.66 &0.8 &0.58 &0.73 &0.48 &0.68 \\\cmidrule{1-9}
20 &0.05 &0.12 &0.62 &0.77 &0.48 &0.65 &0.44 &0.65 \\\cmidrule{1-9}
25 &0.04 &0.12 &0.62 &0.76 &0.42 &0.6 &0.47 &0.67 \\\cmidrule{1-9}
50 &0.04 &0.1 &0.54 &0.7 &0.31 &0.47 &0.44 &0.65 \\\cmidrule{1-9}
& & & & & & & & \\
{\textbf{3-BP}} & & & & & & & & \\
\textbf{Num\_Steps}
% \multirow{\textbf{Num\_Steps}}
&\multicolumn{2}{c}{\textbf{ICSWM +D - C}} &\multicolumn{2}{c}{\textbf{ICSWM}} &\multicolumn{2}{c}{\textbf{ICSWM + D}} &\multicolumn{2}{c}{\textbf{CSWM}} \\\cmidrule{2-9}
&\textbf{Hits @ 1} &\textbf{MRR} &\textbf{Hits@1} &\textbf{MRR} &\textbf{Hits @ 1} &\textbf{MRR} &\textbf{Hits @ 1} &\textbf{MRR} \\\cmidrule{1-9}
1 &0.48 &0.6 &0.99 &1 &0.99 &0.99 &0.99 &0.99 \\\cmidrule{1-9}
5 &0.05 &0.11 &0.83 &0.9 &0.65 &0.77 &0.95 &0.97 \\\cmidrule{1-9}
10 &0.01 &0.02 &0.4 &0.56 &0.27 &0.41 &0.79 &0.87 \\\cmidrule{1-9}
\bottomrule
\end{tabular}
\end{table*}

\begin{table*}[!htp]\centering
\caption{Standard Error}\label{tab: }
\scriptsize
\begin{tabular}{lrrrrrrrrr}\toprule
& & & & & & & & \\
{\textbf{2-BS}} & & & & & & & & \\
\textbf{Num\_Steps}
% \multirow{\textbf{Num\_Steps}}
&\multicolumn{2}{c}{\textbf{ICSWM +D - C}} &\multicolumn{2}{c}{\textbf{ICSWM}} &\multicolumn{2}{c}{\textbf{ICSWM + D}} &\multicolumn{2}{c}{\textbf{CSWM}} \\\cmidrule{2-9}
&\textbf{Hits @ 1} &\textbf{MRR} &\textbf{Hits@1} &\textbf{MRR} &\textbf{Hits @ 1} &\textbf{MRR} &\textbf{Hits @ 1} &\textbf{MRR} \\\cmidrule{1-9}
1 &0.033 &0.047 &0.165 &0.115 &0.075 &0.046 &0.033 &0.031 \\\cmidrule{1-9}
5 &0.006 &0.01 &0.177 &0.134 &0.153 &0.108 &0.012 &0.015 \\\cmidrule{1-9}
10 &0.003 &0.005 &0.222 &0.192 &0.144 &0.097 &0.015 &0.013 \\\cmidrule{1-9}
15 &0.007 &0.006 &0.23 &0.2 &0.163 &0.111 &0.01 &0.007 \\\cmidrule{1-9}
20 &0.009 &0.009 &0.223 &0.207 &0.183 &0.134 &0.01 &0.009 \\\cmidrule{1-9}
25 &0 &0.002 &0.214 &0.214 &0.189 &0.145 &0.003 &0.003 \\\cmidrule{1-9}
50 &0.003 &0.002 &0.13 &0.163 &0.23 &0.192 &0.015 &0.009 \\\cmidrule{1-9}
{\textbf{3-BS}} & & & & & & & & \\
\textbf{Num\_Steps}
% \multirow{\textbf{Num\_Steps}}
&\multicolumn{2}{c}{\textbf{ICSWM +D - C}} &\multicolumn{2}{c}{\textbf{ICSWM}} &\multicolumn{2}{c}{\textbf{ICSWM + D}} &\multicolumn{2}{c}{\textbf{CSWM}} \\\cmidrule{2-9}
&\textbf{Hits @ 1} &\textbf{MRR} &\textbf{Hits@1} &\textbf{MRR} &\textbf{Hits @ 1} &\textbf{MRR} &\textbf{Hits @ 1} &\textbf{MRR} \\\cmidrule{1-9}
1 &0.118 &0.151 &0.058 &0.03 &0.019 &0.01 &0.039 &0.027 \\\cmidrule{1-9}
5 &0.05 &0.071 &0.049 &0.025 &0.079 &0.047 &0.033 &0.02 \\\cmidrule{1-9}
10 &0.027 &0.036 &0.054 &0.029 &0.125 &0.081 &0.049 &0.033 \\\cmidrule{1-9}
15 &0.017 &0.019 &0.05 &0.031 &0.15 &0.109 &0.009 &0.004 \\\cmidrule{1-9}
20 &0.003 &0.01 &0.04 &0.024 &0.156 &0.119 &0.059 &0.039 \\\cmidrule{1-9}
25 &0 &0.008 &0.03 &0.027 &0.186 &0.143 &0.009 &0.006 \\\cmidrule{1-9}
50 &0.003 &0.001 &0.094 &0.094 &0.23 &0.187 &0.037 &0.024 \\\cmidrule{1-9}
{\textbf{3-BSC}} & & & & & & & & \\
\textbf{Num\_Steps}
% \multirow{\textbf{Num\_Steps}}
&\multicolumn{2}{c}{\textbf{ICSWM +D - C}} &\multicolumn{2}{c}{\textbf{ICSWM}} &\multicolumn{2}{c}{\textbf{ICSWM + D}} &\multicolumn{2}{c}{\textbf{CSWM}} \\\cmidrule{2-9}
&\textbf{Hits @ 1} &\textbf{MRR} &\textbf{Hits@1} &\textbf{MRR} &\textbf{Hits @ 1} &\textbf{MRR} &\textbf{Hits @ 1} &\textbf{MRR} \\\cmidrule{1-9}
1 &0.062 &0.116 &0.081 &0.042 &0.003 &0.002 &0.05 &0.038 \\\cmidrule{1-9}
5 &0.064 &0.086 &0.062 &0.026 &0.02 &0.01 &0.088 &0.062 \\\cmidrule{1-9}
10 &0.033 &0.055 &0.038 &0.036 &0.007 &0.007 &0.035 &0.031 \\\cmidrule{1-9}
15 &0.038 &0.067 &0.069 &0.057 &0.043 &0.038 &0.038 &0.035 \\\cmidrule{1-9}
20 &0.027 &0.051 &0.11 &0.085 &0.017 &0.018 &0.074 &0.052 \\\cmidrule{1-9}
25 &0.02 &0.041 &0.128 &0.103 &0.053 &0.045 &0.064 &0.049 \\\cmidrule{1-9}
50 &0.018 &0.035 &0.163 &0.14 &0.043 &0.05 &0.062 &0.05 \\\cmidrule{1-9}
{\textbf{3-BP}} & & & & & & & & \\
\textbf{Num\_Steps}
% \multirow{\textbf{Num\_Steps}}
&\multicolumn{2}{c}{\textbf{ICSWM +D - C}} &\multicolumn{2}{c}{\textbf{ICSWM}} &\multicolumn{2}{c}{\textbf{ICSWM + D}} &\multicolumn{2}{c}{\textbf{CSWM}} \\\cmidrule{2-9}
&\textbf{Hits @ 1} &\textbf{MRR} &\textbf{Hits@1} &\textbf{MRR} &\textbf{Hits @ 1} &\textbf{MRR} &\textbf{Hits @ 1} &\textbf{MRR} \\\cmidrule{1-9}
1 &0.11 &0.1 &0 &0 &0.01 &0.01 &0.01 &0 \\\cmidrule{1-9}
5 &0.02 &0.04 &0 &0 &0.11 &0.08 &0.02 &0.01 \\\cmidrule{1-9}
10 &0 &0 &0 &0 &0.07 &0.07 &0.02 &0.02 \\\cmidrule{1-9}
\bottomrule
\end{tabular}
\end{table*}

\end{enumerate}

\end{document}